\documentclass[lettersize,journal]{IEEEtran}
\usepackage{amsmath,amsfonts}
\usepackage{algorithmic}
\usepackage{algorithm}
\usepackage{array}
\usepackage[caption=false,font=normalsize,labelfont=sf,textfont=sf]{subfig}
\usepackage{textcomp}
\usepackage{stfloats}
\usepackage{xurl}
\usepackage{verbatim}
\usepackage{graphicx}
\usepackage{cite}
\usepackage{orcidlink}

\newcommand\bluecite[1]{\textcolor{blue}{\cite{#1}}}

\begin{document}
\title{Enhancing DeepLabV3+ to Fuse Aerial and Satellite Images for Semantic Segmentation 
\thanks{This project has been funded by the Ministry of Europe and Foreign Affairs (MEAE), the Ministry of Higher Education, Research (MESR) and  the Ministry of Higher Education, Scientific Research and Innovation (MESRSI), under the framework of the Franco-Moroccan bilateral program PHC TOUBKAL Toubkal/21/121 2023, with Grant number 45942UG.}}

\author{
    Anas {BERKA}$^{1,2,*}$\orcidlink{0000-0002-6282-3972}, Mohamed {EL HAJJI}$^{1}$\orcidlink{0000-0002-0327-8249}, Raphael {CANALS}$^{3}$\orcidlink{0000-0001-9100-7539}, Youssef {ES-SAADY}$^{1}$\orcidlink{0000-0002-4934-2322}, Adel {HAFIANE}$^{2}$\orcidlink{0000-0003-3185-9996} \\
    
    \IEEEauthorblockA{$^{1}$IRF-SIC Laboratory, Ibnou Zohr University, BP 8106—Cite Dakhla, Agadir 80000, Morocco.} \\
    \IEEEauthorblockA{$^{2}$INSA CVL, University of Orleans, PRISME Laboratory, EA 4229, Bourges 18022, France} \\
    \IEEEauthorblockA{$^{3}$University of Orleans, Polytech, PRISME Laboratory EA 4229, Orleans, 45072, France.} \\
    \thanks{$^{*}$ Corresponding author's email address: \\ anas.berka@insa-cvl.fr (Anas BERKA \orcidlink{0000-0002-6282-3972})}
}
\maketitle

\begin{abstract}
    Aerial and satellite imagery are inherently complementary remote sensing sources, offering high-resolution detail alongside expansive spatial coverage. However, the use of these sources for land cover segmentation introduces several challenges, prompting the development of a variety of segmentation methods. Among these approaches, the DeepLabV3+ architecture is considered as a promising approach in the field of single-source image segmentation. However, despite its reliable results for segmentation, there is still a need to increase its robustness and improve its performance. This is particularly crucial for multimodal image segmentation, where the fusion of diverse types of information is essential. 
    An interesting approach involves enhancing this architectural framework through the integration of novel components and the modification of certain internal processes.
    In this paper, we enhance the DeepLabV3+ architecture by introducing a new transposed conventional layers block for upsampling a second entry to fuse it with high level features. This block is designed to amplify and integrate information from satellite images, thereby enriching the segmentation process through fusion with aerial images.  
    For experiments, we used the LandCover.ai (Land Cover from Aerial Imagery) dataset for aerial images, alongside the corresponding dataset sourced from Sentinel 2 data.
    Through the fusion of both sources, the mean Intersection over Union (mIoU) achieved a total mIoU of 84.91\% without data augmentation.    
\end{abstract}


\section{Introduction}

The field of remote sensing (RS) has undergone significant advancements through the utilization of multiple available imagery sources and their application in various domains \bluecite{alvarez-vanhard_uav_2021}. The availability of RS data from a variety of sources, including satellite, aerial and drone imagery, has increased significantly in recent years \bluecite{noauthor_directive_nodate,indigeo_projets_nodate}. The combination of remote sensing data using the computer vision and artificial intelligence (AI) allows the extraction and analysis of a wide range of information. This approach has been used in a variety of applications, including scene classification \bluecite{Rupasinghe_Simic_Milas_Arend_Simonson_Mayer_Mackey_2019}, land monitoring \bluecite{isprs-archives-XLII-2-W13-297-2019} and spatially semantic segmentation \bluecite{Calleja,Lv_Shen_Lv_Li_Shi_Zhang_2023}. Consequently, serval methods have been developed, including various new versions of classic deep learning models such as Unet \bluecite{ronneberger_u-net_2015}, Segnet \bluecite{badrinarayanan_segnet_2017}, and DeepLab \bluecite{Li_2020,Mo_Wu_Yang_Liu_Liao_2022}. Nevertheless, there is a continued necessity to enhance the robustness and performance of existing models. 

In remote sensing, semantic segmentation techniques have focused on higher dimensional representations of complex scenes, allowing significant information to be extracted from substantial amounts of data \bluecite{Thisanke_Deshan_Chamith_Seneviratne_Vidanaarachchi_Herath_2023,Wieland_Martinis_Kiefl_Gstaiger_2023}. It involves assigning a class label to each pixel of an image, allowing for precise analysis of these complex scenes. 
Recent research has focused on using encoder-decoder architectures to improve the performance of semantic segmentation \bluecite{wang_unetformer_2022,ovi_deeptrinet_2023,dimitrovski_u-net_2024}. For example, DeepLab has demonstrated remarkable success in handling challenging scenes with complex structures and varying scales \bluecite{Anilkumar_Venugopal_Maddikunta_Gadekallu_Al_Rasheed_Abbas_Soufiene_2023}. It is one of the most effective semantic segmentation architecture currently available. The efficacy of DeepLab has been demonstrated by numerous studies \bluecite{boguszewski_landcoverai_2022, Wang_Wang_Wu_Chen_2022, Ovi_Mosharrof_Bashree_Islam_Islam_2023}. Initially developed by \bluecite{Chen_Papandreou_Kokkinos_Murphy_Yuille_2016}, it has undergone continuous refinement and enhancement over time. The last official version DeepLabV3+ \bluecite{Chen_Zhu_Papandreou_Schroff_Adam_2018} is based on an encoder-decoder architecture. It incorporates a range of methods and techniques, including atrous convolution (AC), atrous spatial pyramid pooling (ASPP), and fully connected conditional random fields (CRFs). These techniques contribute to providing a more refined representation of the semantics of the input by enhancing the accuracy and precision. In addition, DeepLabV3+ employs a pixel-wise classification approach, whereby semantic labels are assigned to each pixel in the input image. This enables the precise delineation of objects and features within the scene. Despite the considerable capabilities of semantic segmentation models, the fusion of low-resolution satellite data with high-resolution aerial images presents a significant challenge. The DeepLabV3+ model was designed primarily for single-source image segmentation tasks and has demonstrated promising results while fusing multiple sources of images with the same resolution. The complementary nature of data sources, such as high-resolution aerial imagery and broader coverage satellite data, presents both opportunities and challenges for researchers seeking accurate and comprehensive segmentation results \bluecite{ghamisi_multisource_2018}. 

Several studies have explored the use of aerial images and satellite data for semantic segmentation tasks. High-resolution aerial images provides detailed information about land cover, but its coverage is often limited \bluecite{Shahi_Xu_Neupane_Guo_2023}. In contrast, satellite data, offers broader spatial coverage but with lower resolution \bluecite{Panagiotopoulou_Grammatikopoulos_Kalousi_Charou_2021}. In fact, the fusion of aerial and satellite data presents a compelling opportunity for enhancing semantic segmentation, given the inherent complementarity of these data sources. Aerial images typically provide higher spatial resolution, which is crucial for capturing fine details, but they are often limited to RGB bands and lack broader spectral coverage. In contrast, satellite data, such as those from Sentinel, provide multispectral data that can enhance the detection of various land cover types, vegetation health, and urban materials, thus contributing to a richer segmentation. A key challenge lies in effectively fusing these complementary data sources to leverage their strengths for improved semantic segmentation. Existing approaches often involve separate processing of aerial and satellite data, followed by complex fusion strategies \bluecite{Kakooei_Baleghi_2017, Miranda_Pina_Heleno_Vieira_Mora_E_G_R_Schaefer_2020, Heidarian_Dehkordi_Pelgrum_Meersmans_2022}. 

In this context, we propose a novel approach for fusing low-resolution satellite data and high-resolution aerial images within a single DeepLabV3+ model for semantic segmentation. To this end, we take a step back and rethink the following question: Can we exploit the potential of DeepLabV3+ and develop a single model that is both effective and applicable for fusion between satellite data and aerial images for semantic segmentation? To address this question, we propose a novel upsampling block designed to inject information extracted from satellite data into the encoded features of the aerial image during the decoding phase. This leverages the complementary nature of the data sources: high-resolution detail from aerial imagery and rich information from satellite multispectral imagery. We hypothesize that this injection improves model performance and segmentation accuracy. To our knowledge, this is the first study to perform such fusion using a single DeepLabV3+ model. Our Dual-Input Fusion network for aerial and satellite images based on an enhanced DeepLabV3+ (DIFD) architecture incorporates a new transposed convolutional layers block specifically designed for satellite data injection. This block utilizes deconvolution layers to refine the upsampling process, leading to improved reconstruction accuracy in the segmentation map. 
We further enhanced DeepLabV3+ by replacing bilinear upsampling with a weighted upsampling module in the decoder phase, alongside other refinements.  

To validate our approach, we conducted experiments comparing four upsampling methods for satellite data injection within DIFD: bilinear, nearest neighbor, super-resolution with pixel shuffle, and deconvolution. This paper makes a significant contribution by introducing an innovative enhancement to the DeepLabV3+ model, specifically designed for fusing satellite imagery and aerial images. This fusion enables improved performance in land cover semantic segmentation tasks. The model combines high-resolution aerial imagery and low-resolution satellite data in a single deep learning model. 

This paper is organized as follows: the first section provides an introduction to the paper. The second section presents the related work. The third section presents a description of the methods used, while the fourth section provides precision on the exact flow for the experimentation. The results are analyzed in the fifth section, and conclusions are drawn in the final section.

\section{Related work}

High-resolution aerial imagery excels at capturing detailed land cover information, providing valuable insights into features such as vegetation types and building structures \bluecite{Shahi_Xu_Neupane_Guo_2023}. However, their coverage is often limited to specific areas of interest due to acquisition costs and flight scheduling constraints. In contrast, satellite imagery provides broader spatial coverage, allowing analysis of large geographical areas. However, satellite imagery typically has a lower spatial resolution, making it difficult to distinguish fine-grained details such as individual trees or small buildings or roads \bluecite{Ghandorh_Boulila_Masood_Koubaa_Ahmed_Ahmad_2022}. Fusing these complementary data sources is a major challenge. The key obstacle is to effectively combine the high-resolution detail of aerial images with the multispectral information coverage of satellite data. This requires appropriate techniques to deal with differences in resolution and information content between the two sources. For example, \bluecite{Ghandorh_Boulila_Masood_Koubaa_Ahmed_Ahmad_2022} used Spot 7 satellite data (with spatial resolution of 6 m/pixel in the multispectral channels and 1.5 m/pixel in the panchromatic band) to identify roads in selected regions. Their approach used an encoder with attention maps followed by two separate blocks for pixel segmentation and edge detection, achieving accurate edge detection even in complex backgrounds. However, the Sentinel-2, another popular and more publicly available choice for land cover analysis, offers a lower spatial resolution of 10 m/pixel and lacks a panchromatic band \bluecite{Panagiotopoulou_Grammatikopoulos_Kalousi_Charou_2021}. This highlights the trade-offs between resolution and coverage inherent in different satellite sensors.

Despite the challenges posed by class similarity, class diversity, and texture complexity \bluecite{Shahi_Xu_Neupane_Guo_2023}, aerial images remains valuable for land cover segmentation. In \bluecite{Kerkech_Hafiane_Canals_2020}, the authors propose the VddNet architecture to segment and detect diseases in grapevines from UAV images by fusing RGB input with infrared (IR) and depth information. Similarly, \bluecite{He_Li_Yang_Zeng_Li_Zhu_2022} used deep learning for land cover segmentation, with a specific focus on landslide extraction. Their approach fused spectral and topographic features derived from drone photogrammetry to achieve accurate segmentation. \bluecite{Akcay_Kinaci_Avsar_Aydar_2021} proposed a dual-stream DeepLabV3+ architecture to address these challenges. This approach processes information from a drone raster input consisting of RGB channels and a secondary input containing a normalized digital surface model (nDSM), infrared (IR) and normalized difference vegetation index (NDVI). By using dual encoders for these separate inputs and concatenating the encoded features in the decoder phase, the model achieves improved performance compared to a single-encoder approach that processes all information simultaneously.  The examples highlight the potential of deep learning models for land cover segmentation, particularly when leveraging the strengths of both aerial and satellite imagery. However, overcoming the inherent differences in resolution and information content between these sources remains a significant challenge in the field.

DeepLabV3+ \bluecite{Chen_Zhu_Papandreou_Schroff_Adam_2018}, is a well-established and versatile architecture for semantic segmentation tasks. In a manner analogous to its predecessor, DeepLabV3 \bluecite{chen_rethinking_2017}, DeepLabV3+ incorporates the concept of an encoder-decoder into its architectural framework. Additionally, the modified Xception backbone employs separable convolutions for the purpose of efficient feature extraction. This design has proven remarkably effective on various datasets, including the PASCAL VOC 2012 benchmark \bluecite{pascal_voc_2012}. Researchers have continuously improved DeepLabV3+ by exploring alternative components within its architecture. For example, \bluecite{Chen_Collins_Zhu_Papandreou_Zoph_Schroff_Adam_Shlens_2018} compared the impact of using the original Atrous Spatial Pyramid Pooling (ASPP) block, a key component for capturing multiscale information, with a Dense Prediction Cell (DPC). Their analysis helped to refine the feature extraction process by identifying optimal connections between atrous convolutions within the DPC block. In \bluecite{Kamann_2020_CVPR}, the authors conducted a benchmark analysis to evaluate the robustness of DeepLabV3+ configurations. This analysis provided a more in-depth comparison of different component options and confirmed the importance of atrous convolution as a core component for DeepLabV3+. However, the choice between ASPP and DPC depends on the specific data characteristics. While DPC may perform better on clean data, ASPP excels when dealing with corrupted data containing noise or defocus blur. Furthermore, \bluecite{Wang_Wang_Wu_Chen_2022} proposed an improvement to the original DeepLabV3+ model by replacing the Xception backbone with a Res2Net architecture and incorporating a multi-loss function. This function targets feature information from different network layers, allowing the model to use shallower information during backpropagation to adjust parameters more effectively.

The decoding phase in DeepLabV3+ uses a simple interpolation method for upsampling. This process enlarges the final slices to produce the final prediction that matches the dimensions of the input image. However, a challenge arises for classes representing small regions in land cover tasks. For example, the LandCover.ai dataset \bluecite{boguszewski_landcoverai_2022} contains building and road classes represented by small areas or thin lines, particularly in urban environments. Without appropriate upsampling techniques, information for these classes may be lost in the process. Consequently, there is a need for improved upsampling methods within the model to capture these fine details, especially when dealing with data sources of different spatial resolutions. The standard bilinear interpolation upsampling method in DeepLabV3+ may encounter difficulties in preserving the details of small classes (e.g., buildings, roads) during land cover segmentation. This is particularly evident when dealing with data sources of different spatial resolutions, such as high-resolution aerial images and lower-resolution satellite data. Several techniques have been developed to address the upsampling challenges previously outlined. Two prominent approaches are pansharpening and super-resolution. Pansharpening is a technique that aims to enhance the resolution or information content of the multispectral bands (e.g., red, green, blue, near-infrared) in satellite imagery by incorporating information from the higher-resolution panchromatic band \bluecite{loncan_hyperspectral_2015,neale_pan-sharpening_2016}. This can be particularly useful when fusing images with different spatial resolutions, as demonstrated by \bluecite{Ciotola_Vitale_Mazza_Poggi_Scarpa_2022}. In addition,  techniques such as ESRGAN utilize Generative Adversarial Networks (GANs) in combination with pansharpening to achieve image enhancement \bluecite{Wang_Yu_Wu_Gu_Liu_Dong_Qiao_Loy_2019}. Super-resolution is an approach that aims to reconstruct a high spatial resolution image from a low spatial resolution input, based on traditional super-resolution techniques \bluecite{Dong_Loy_He_Tang_2016}. This method is designed to achieve this reconstruction despite potential limitations in the number of channels and initial spatial resolution. A common approach within super-resolution methodologies for upsampling tensors is pixel shuffling, which reorganizes depth information to create a new higher-resolution output \bluecite{Chatterjee_Chu_2020}.  
Alternatively, transpose convolution offers another approach for upsampling feature maps within the decoder path of a deep learning model. It was introduced in the U-Net architecture \bluecite{ronneberger_u-net_2015} in order to upsample the input in the expansion path. It was then further demonstrated in \bluecite{dimitrovski_u-net_2024} that the ensemble method using U-net as a base with advance techniques for processing the data can achieve a higher results and precision using aerial images input only. By employing the upsampling or pre-processing techniques, researchers may potentially enhance the accuracy of segmentation for small classes in land cover tasks, particularly when dealing with multi-source imagery of varying resolutions.

\section{Methods}

This section presents a detailed description of the methodology used in our work, including data pre-processing and the proposed architecture. Our approach enhances not only DeepLabV3+ but also land cover segmentation methodologies by integrating various remote sensing data sources. The model integrates high spatial resolution aerial images (RGB) and low spatial resolution satellite data that had higher spectral resolution into a unified deep learning model.

\subsection{Architecture}
\begin{figure*}[htb]
    \centering
    \includegraphics[width=\textwidth]{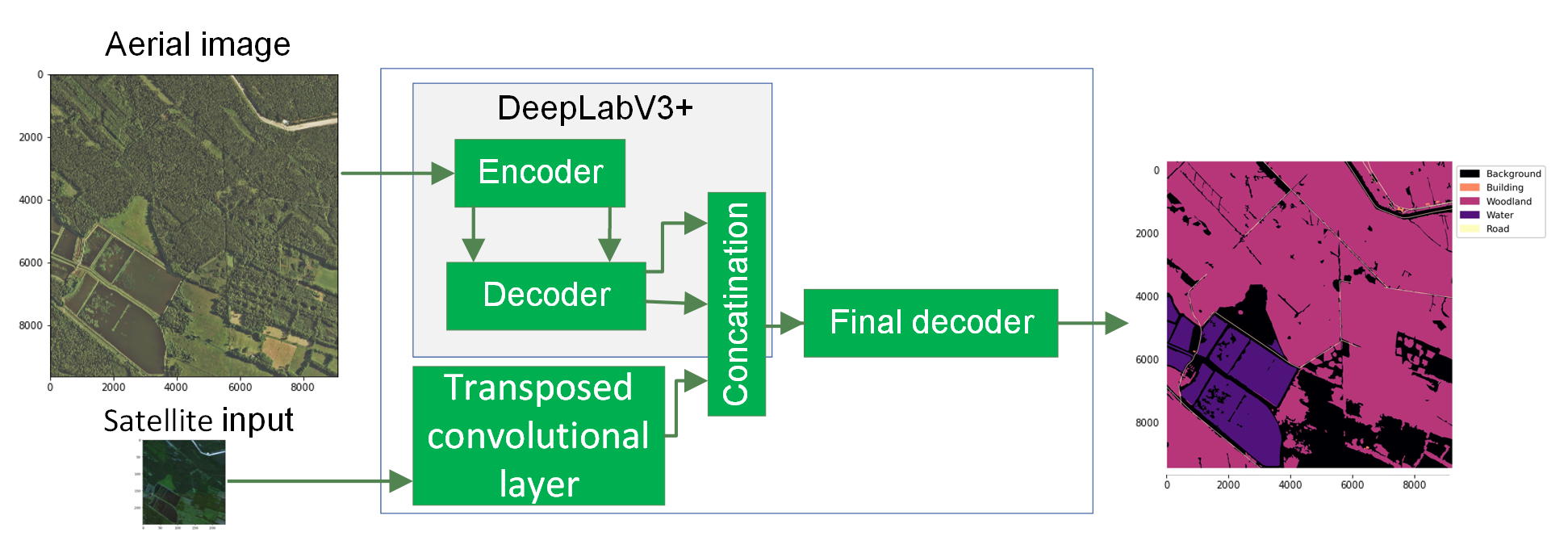}
    \caption{Overview of the proposed architecture.}
    \label{fig:overview}
\end{figure*}

An overview of the proposed architecture is illustrated in Fig.\ref{fig:overview}, which requires two inputs: aerial images and satellite data. These inputs are processed separately before being fused to generate the final segmented map. Our proposed architecture for enhancing DeepLabV3+ features a new deconvolution block for the second input, while also targeting each component of the base model with an enhancement to further improve the performance.

\begin{figure*}[htb]
    \centering
    \includegraphics[width=\textwidth]{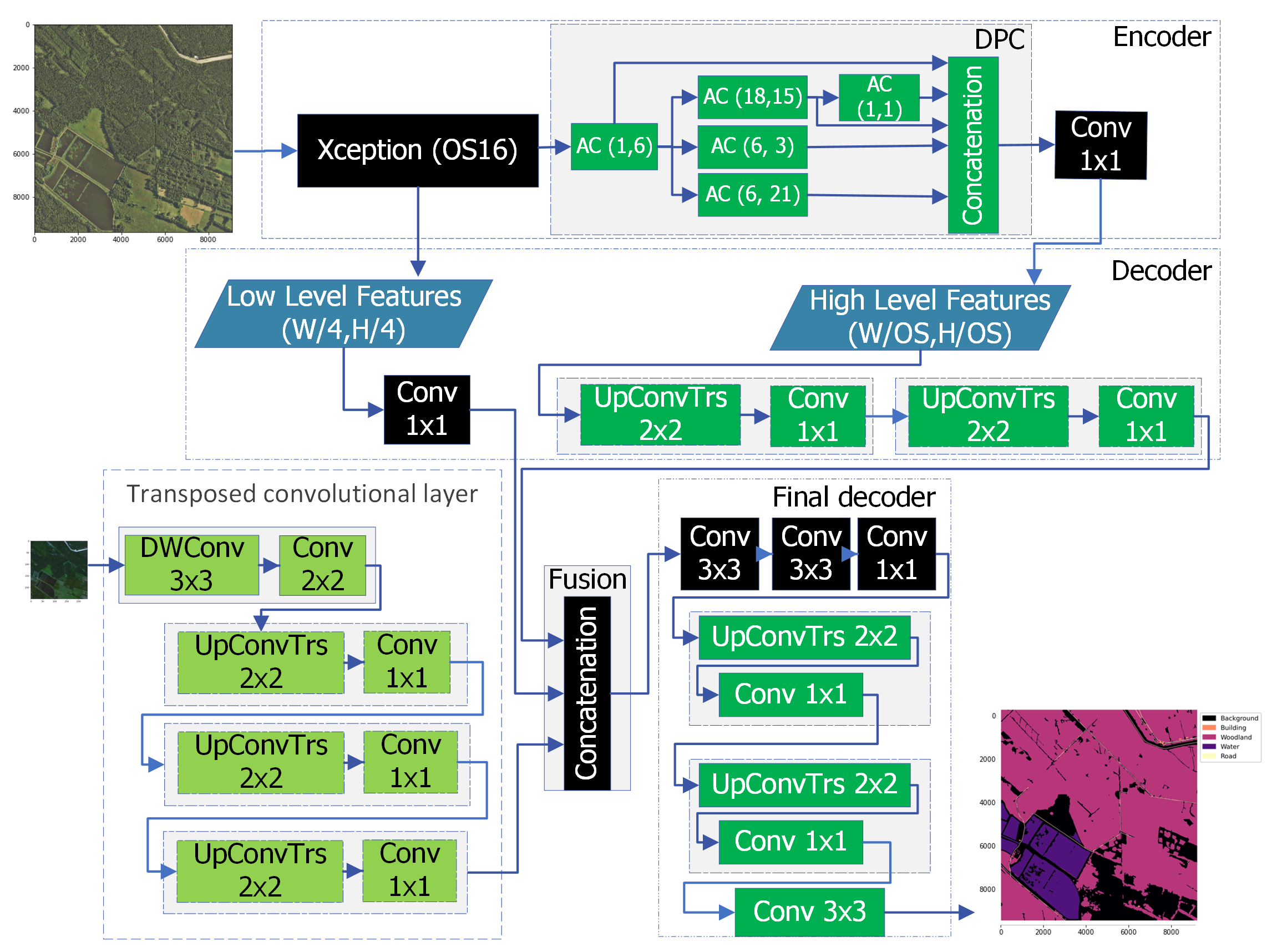}
    \caption{Our proposed architecture Dual-Input Fusion network for aerial images and satellite raster data based on an enhanced DeepLabV3+ (DIFD)}
    \label{fig:DIFD_F5}
\end{figure*}

For a more detailed illustration, Fig.\ref{fig:DIFD_F5} presents the Dual-Input Fusion network based on the DeepLabV3+ (DIFD) architecture. This network was built upon DeepLabV3+, incorporating several enhancements specifically designed for land cover segmentation with multi-source remote sensing data. The modifications and additions to the base DeepLabV3+ architecture are highlighted in Fig.\ref{fig:DIFD_F5} for clarity. The block that has been added for the purpose of information fusion is colored yellow, enhancements to existing blocks are colored green, and the original DeepLabV3+ blocks remain black.

\subsubsection{Encoder}
The encoder in the proposed architecture is similar to the DeepLabV3+ encoder. It consists of two main components: a feature extractor and a refinement block for processing the high-level features. The feature extractor utilizes a modified Xception backbone with an output stride (OS) of 16 \bluecite{wightman2023xception} pre-trained on the ImageNet1K dataset \bluecite{imagenet15russakovsky}. 
The pre-trained model extracts informative low-level features (LLF) from the second internal block of Xception while processing the input imagery. For high-level features (HLF) extracted at the end of the backbone, instead of the standard ASPP block, it employs a the DPC block followed by a pointwise convolution 1x1 to refine the encoded output of the feature extractor \bluecite{Chen_Collins_Zhu_Papandreou_Zoph_Schroff_Adam_Shlens_2018}. This refinement step aims to generate richer feature representations that are more suitable for the subsequent decoding stages. 
This encoding process can be formulated as Eq.\eqref{Eq:encoder}:
\begin{equation}
    \text{(LLF,HLF)}=\text{Encoder}({input_1})
    \label{Eq:encoder}
\end{equation}
where $\text{HLF} \in \mathbb{R}^{m \times m \times 2n}$ and $\text{LLF} \in \mathbb{R}^{n \times n \times n}$ are the output of the encoder with $m=32$ and $n=128$. Let $\boldsymbol{input_1} \in \mathbb{R}^{k \times k \times C_1}$ where $\boldsymbol{input_1}$ is a cropped aerial input image, $C_1 = 3$ number of channels (RGB) and $\boldsymbol{k}=512$ the size of the image. $\text{Encoder}$ denotes the encoder block that encode the input to a HLF and LLF.
It uses the Xception (OS16) model mentioned above for feature extraction of HLF and the output of the backbone undergoes more refinement within the DPC, using its atrous convolutions blocks (AC) with various rates as mentioned in Fig.\ref{fig:DIFD_F5} followed by a concatenation and point wise convolution 1x1.

\subsubsection{Decoder for Aerial Imagery (Decoder)}


The Decoder module processes encoded information from aerial images, using low-level features (LLF) to capture global scene characteristics and high-level features (HLF) to provide more detailed, localized information. To improve the retrieval of low-level features during decoding, we employed pointwise 1x1 convolutions with an Exponential Linear Unit (ELU) activation function, replacing ReLU activation utilized in DeepLabV3+. This modification was made on the basis of the evidence that ELU is better suited to handle negative inputs, as demonstrated by \bluecite{clevert_fast_2015}, who showed that ELU exhibited superior performance compared to ReLU. Our own experimental results confirmed that this adjustment led to enhanced overall segmentation performance.
This can be formulated as Eq.\eqref{Eq:decoder1L}:
\begin{equation}
    \text{llf1}=\boldsymbol{W} \cdot \text{LLF}+\boldsymbol{b}
    \label{Eq:decoder1L}
\end{equation}
Let $\text{llf1} \in \mathbb{R}^{n \times n \times 48}$, it is the output of the LLF been processed by a single convolution 1x1 with $W$ and $b$ as the weights and biases.

To further refining the high-level features, the conventional approach of simple bilinear upsampling is replaced with a novel block $UpConvT$ \bluecite{Berka_2014}, which employs weighted upsampling. The upsampling block comprises two iterations of a deconvolution that uses $ConvolutionTranspose2D$ from Pytorch that we noted as $UpConvTrs$ with a kernal of 2x2 followed by a pointwise convolution 1x1, batch normalization (BN), and ELU activation. 
This methodology facilitates the model's capacity to learn more effective scaling factors for the extracted features during the upsampling process. 
The process can be formulated as Eq.\eqref{Eq:decoder1H}:
\begin{multline}
    \text{hlf}=f_1\circ f_2(\text{HLF})\\
    \text{ where} f_i = \text{ELU}(\text{BN}(\text{Conv}_{1\times1}(\text{UpConvTrs}))) \text{ and } i\in \left [ 1, 2 \right ]
    \label{Eq:decoder1H}
\end{multline}
Let $\text{hlf} \in \mathbb{R}^{n \times n \times 256}$, it is the output of the HLF after been processed by the upsampling block, in which $f$ denoted the process for deconvolution using $UpConvTrs$ 2x2 followed by the  pointwise convolution 1x1 then BN and ELU. 
$f$ were applied two times in order to have the targeted tensor size.
Hence, $hlf$ and $llf1$ and  are the outputs of the decoder block.

\subsubsection{Upsampling Satellite data}


In order to effectively incorporate information from satellite data, transposed convolutional layers are employed for the processing of the satellite data. Transposed convolutional layers are selected for their aptitude for upsampling, which enables an enhancement in the spatial resolution of feature maps, increasing the depth, the width, and height of the output. The upscaling process begins with a sub-block consisting of a depthwise convolution noted in Fig.\ref{fig:DIFD_F5} as $DWConv$ 3x3, followed by a standard convolution 2x2 and batch normalization, this sub-block serve to augment the feature depth. 
In a comparable manner to the upsampling strategy employed for high-level features in Decoder block, in this block (transposed convolutional layer) it employs a greater number of layers due to the potential for a greater discrepancy between the input size and the desired output size. Thus, it is composed of three repetitions of the $UpConvTrs$ 2x2 with a convolution 2x2, BN and ELU activation sub-block, which serves to facilitate effective upsampling. 
The transposed convolutional layers (TCL) block can be represented as in Eq.\eqref{Eq:UpConvT}.


\begin{multline}\label{Eq:UpConvT}
    \text{llf2} = \text{TCL}_{\text{UpConvT}}(\boldsymbol{input_2}) \\
    = f_1 \circ f_2 \circ f_3 (\text{BN}(\text{Conv}_{2\times2}(\text{DWConv}(\boldsymbol{input_2})))) \\
    \text{ where TCL stands for transposed convolutional layers}
\end{multline}

Let $\boldsymbol{input_2} \in \mathbb{R}^{26 \times 26 \times C2}$ the corresponding cropped raster input from the satellite. 
$C2$ is the number of bands to be used.
The upsampled output is considered a second set of low-level features, providing complementary information to those extracted from the aerial imagery. Using this decoding method that uses multiple blocks of UpConvT, we alternatively named this version of our DIFD as DIFD\_UpConvT.

\label{Variation_of_DFID}
The DIFD architecture offers other variations for the upsampling method for the second input. In the DIFD\_UpNearest method we employed nearest neighbor interpolation to preserve the original values by upscaling the satellite raster directly to the size of 128 by 128 as in Eq.\eqref{eq:UpNearest}. 
\begin{equation}
    \begin{aligned}
        \text{llf2}_{\text{UpNearest}} &= \text{Interpolation}_{\text{Nearest}}(\boldsymbol{input_2})\\
    \end{aligned}
    \label{eq:UpNearest}
\end{equation}

Whereas the DIFD\_UpBilinear method utilizes bilinear interpolation for more gradual upsampling the satellite input to the target resolution Eq.\eqref{eq:UpBilinear}. 
\begin{equation}
    \begin{aligned}
        \text{llf2}_{\text{UpBilinear}} &= \text{Interpolation}_{\text{Bilinear}}(\boldsymbol{input_2})\\
    \end{aligned}
    \label{eq:UpBilinear}
\end{equation}

In addition to that, DIFD\_UpPS is an inspired method from the super resolution techniques \bluecite{Shi2016}. It uses a triple block of convolution to regulate the spatial resolution and depth of the channel before using the Pixel-Shuffle (PS) operation to upscale the raster to the desired size as in Eq.\eqref{eq:UpPS}. 
\begin{equation}
    \text{llf2}_{\text{UpPS}} = \text{PS}(\text{Conv}^{2}(\text{UpConvTrs}(\boldsymbol{x_2})))
    \label{eq:UpPS}
\end{equation}
PS reorders the input tensor elements from $(\text{batch}, \text{filters}, h, w)$ to $(\text{batch}, \text{filters}/ r^{2}, h\times r, w\times r)$, where $r=8$ is the rescaling factor. The reason for using PS is its ability to incorporate in-depth information from the input into the upsampled result. Each pixel in the output corresponds to the outcome of the convolution performed before the application of Pixel Shuffle. Therefore, the model will learn to adjust the weights of the corresponding filters.

\subsubsection{Feature Concatenation and Final Decoder}
The processed outputs from Decoder (llf1 and hlf: aerial information) and transposed convolutional layers (llf2: satellite information) are concatenated to create a combined feature representation. This fused representation incorporates the strengths of both data sources, potentially leading to improved segmentation accuracy. This fused data is then refined by a final decoder consisting of two convolution 3x3 then a pointwise convolution 1x1 in the first sub-block. Then, two $UpConvT$ blocks for upsampling similar to the approach used above. Finally, a convolution 3x3 to fine-tunes the pixel classifications in the segmented output.

\subsection{Datasets}
\begin{figure*}[htb]
    \centering
    \includegraphics[width=\textwidth]{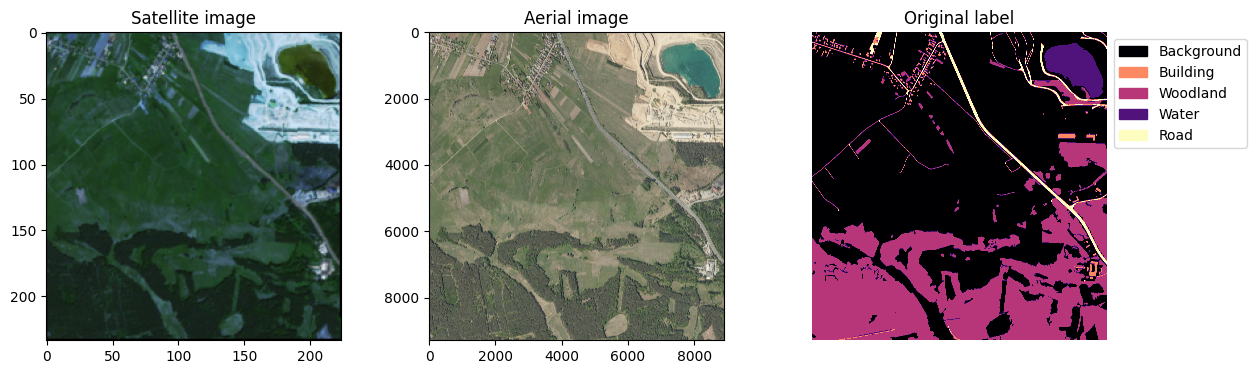}
    \caption{An aerial image from the LandCover.ai dataset, its corresponding satellite image and the ground truth label}
    \label{fig:diff_sat_ai_bp}
\end{figure*}

This study used aerial and satellite data from two sources to analyze the proposed architecture. The aerial data was obtained from the LandCover.ai dataset \bluecite{boguszewski_landcoverai_2022}, which comprises 41 images covering 216.27 km2 of rural areas in Poland. The precise locations of the aerial images, based on geolocation information, were described in detail by \bluecite{boguszewski_landcoverai_2022}. Regarding the satellite data, we extracted only the relevant areas from the original Sentinel-2  raster's using Sentinel-Hub tools \bluecite{sentinel-hub_eo_2023}. Fig. \ref{fig:diff_sat_ai_bp} shows an example of an aerial image with its corresponding cropped satellite image (plotting only RGB bands) and the provided ground truth. 

In order to collect the data, each source of images was collected separately. 
\begin{itemize}
    \item \textbf{Aerial Images}: The images consisted of high-resolution RGB orthophotos. There were 33 orthophotos with a resolution of 25 cm per pixel ( 9000 x 9500 px) and 8 orthophotos with a resolution of 50 cm per pixel (4200 x 4700 px). The images were captured over a period of four years (2015-2018) due to the varying number of flights on which they were based. The orthophotos were stored in GeoTiff format with the EPSG: 2180 spatial reference system \bluecite{boguszewski_landcoverai_2022}.
    \item \textbf{The satellite rasters}: from Sentinel-2 were collected by extracting all bands using the Sentinel-hub API \bluecite{sentinel-hub_eo_2023}. 
    The bands, as described in Table \ref{tab:Ch_Sat}, were provided based on their respective descriptions, but only a selection were used in this work to add various information to the aerial images : 
    \begin{itemize}
        \item \textbf{Original bands only} that we note as $10B$ due to it having 10 channels. This selection uses only the raw bands from satelite which means : B02, B03, B04, B05, B06, B07, B08, B8A, B11 and B12 without any changes.
        \item \textbf{Visual bands only} that we note as $4B$ due to it having only 4 bands : RGB + NIR.
        \item \textbf{Main selected bands} that we note as $7B$ are the 7 bands : NDVI, NDWI, B02E, B03E, B04E, B08 and SCL. We confirmed this selection after various tests using Optuna supposing that the selection of bands is a hyperparameter, and they were the most effective information to our study.
    \end{itemize}
    The raster dimensions were 225 x 238 pixels, they were in a GeoTIFF format and georeferenced with the EPSG:4360 spatial reference system. 
    The retrieval of the satellite images was based on the metadata that had been registered on the aerial images. Thus the satellite images were acquired during the same time frame as the aerial images, which had been captured.
\end{itemize}

\begin{table}[htb]
\centering
\caption{Channels description from the satellite dataset\\ (all provided bands)}
\footnotesize
\begin{tabular}{p{.2cm}p{.4cm}p{4.9cm}p{1.1cm}}
\hline
Ch. & Name & Description                                         & Resolution   \\ \hline
0       & NDVI & Normalized Difference Vegetation Index               & 10m/px  \\ 
1       & NDWI & Normalized Difference Water Index                    & 10m/px  \\ 
2      & B02  & Blue,  492.4 nm (S2A),  492.1 nm (S2B)                & 10m/px  \\ 
3      & B03  & Green,  559.8 nm (S2A),  559.0 nm (S2B)              & 10m/px  \\ 
4      & B04  & Red,  664.6 nm (S2A),  665.0 nm (S2B)                 & 10m/px  \\ 
5      & B05  & Vegetation red edge,  704.1 nm (S2A),  703.8 nm (S2B) & 20m/px  \\ 
6      & B06  & Vegetation red edge,  740.5 nm (S2A),  739.1 nm (S2B) & 20m/px  \\ 
7      & B07  & Vegetation red edge,  782.8 nm (S2A),  779.7 nm (S2B) & 20m/px  \\ 
8      & B08  & NIR,  832.8 nm (S2A),  833.0 nm (S2B)                 & 10m/px  \\ 
9      & B8A  & Narrow NIR,  864.7 nm (S2A),  864.0 nm (S2B)          & 20m/px  \\ 
10      & B11  & SWIR,  1613.7 nm (S2A),  1610.4 nm (S2B)             & 20m/px  \\ 
11      & B12  & SWIR,  2202.4 nm (S2A),  2185.7 nm (S2B)             & 20m/px  \\ 
12      & B02E & Blue,  Enhanced natural color visualization          & 10m/px  \\ 
13      & B03E & Green,  Enhanced natural color visualization         & 10m/px  \\ 
14      & B04E & Red,  Enhanced natural color visualization           & 10m/px  \\ 
15 & SCL & Scene classification data,  based on Sen2Cor processor,  codelist & 10m/px \\ 
16      & CLD  & Cloud probability,  based on Sen2Cor processor       & 10m/px  \\ 
\end{tabular}
\label{tab:Ch_Sat}
\end{table}

To exploit the satellite raster and use all information available including the correct colors of the maps since the original RGB colors were affected by the atmosphere,  we opted to use atmospherically corrected color bands from Sentinel-2 instead of the standard RGB bands.
\begin{figure}[htb]
    \centering
    \includegraphics[width=.48\textwidth]{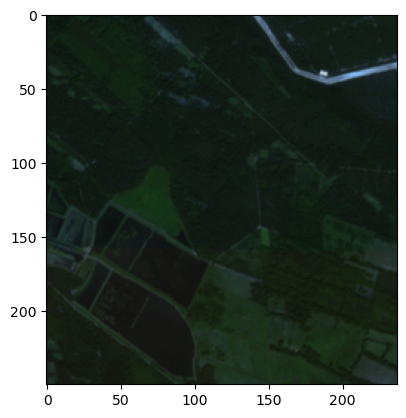}
    \includegraphics[width=.48\textwidth]{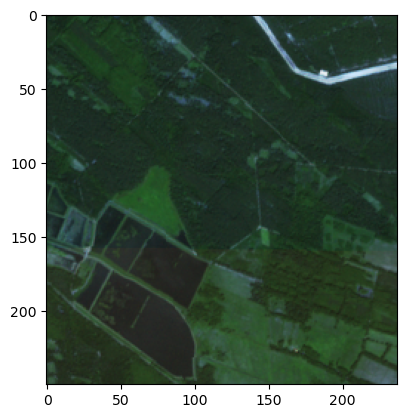}
    \caption{Difference between using the original color RGB from satellite (at the top) and the enhanced RGB color (at the bottom)}
    \label{fig:Sat_diff}
\end{figure}
Fig.\ref{fig:Sat_diff} illustrates the advantage of this choice, as atmospherically corrected bands offer improved clarity compared to the original RGB bands. 
The remaining original bands, which were not utilized, exhibited a lower pixel resolution of 20 m/px. The primary motivation for incorporating the SCL bands was their richer representation of ground truth information from the sentinel's Sen2Cor processor.
Two additional informative bands were incorporated: the Normalized Difference Vegetation Index (NDVI) and the Normalized Difference Water Index (NDWI). These indices are derived directly from bands B03 (Green), B04 (Red), and B08 (Near-Infrared), corresponding to specific wavelengths as detailed in Table \ref{tab:Ch_Sat} (see \bluecite{sentinel-hub_eo_2023} for details). 
The calculations for NDVI and NDWI employ the formulas provided in Eq.\eqref{NDVI}\bluecite{NDVI_1969} and  Eq.\eqref{NDWI}\bluecite{McFEETERS_1996}.

\begin{equation} \text{NDVI} = \frac{B08-B04}{B08+B04} \label{NDVI} \end{equation}

\begin{equation} \text{NDWI} = \frac{B03-B08}{B03+B08} \label{NDWI} \end{equation}

As illustrated in Table \ref{tab:diff_sat_ai}, the Sentinel-2 rasters and aerial images differed in their spatial reference system and resolution. To facilitate effective analysis, we performed spatial alignment by unifying both datasets to the EPSG:2180 reference system. This process involved resampling the Sentinel-2 rasters using the nearest neighbor method from the Rasterio library \bluecite{rasterio}. It was necessary to ensure compatibility between the datasets during the mapping process. This approach to resampling ensures that no information is lost by assigning the nearest neighbor pixel value during the remapping process. 

\begin{table}[htb]
\centering
\caption{Difference between aerial and satellite images }
\begin{tabular}{p{3cm}p{2.5cm}p{1.8cm}}
Specifications             & Aerial images      & Satellite images   \\ \hline
image size        & $9000\times9500$ or $4200\times4700$px & $225\times238$ \\ 
Spatial resolution         & $25$cm   or $50$cm            & $10$m             \\ 
Cropped image size         & $512\times512$         & $26\times26$             \\ 
Channels                   & 3                  & 7                 \\ 
Spatial reference   system & EPSG:2180   & EPSG: 4326         \\ \hline
\end{tabular}
\label{tab:diff_sat_ai}
\end{table}

\subsection{Training and preprocessing Strategy}
\begin{figure}[htb]
    \centering
    \includegraphics[width=\linewidth]{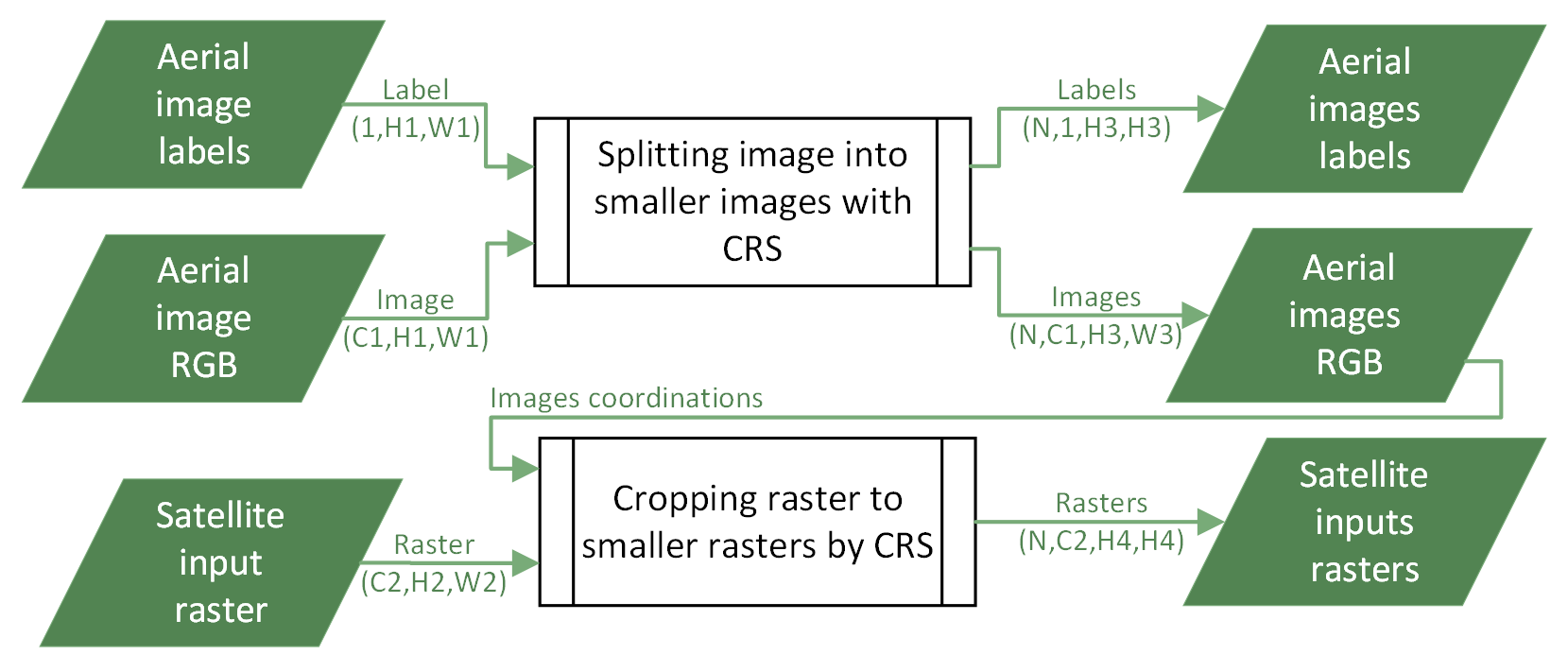}
    \caption{The pre-processing for images}
    \label{fig:preprocess}
\end{figure}

In order to prepare the data for training and inference since the original aerial images' sizes are too large ($9000 \times 9500$) or ($4200 \times 4700$), a slicing for the images is needed. This process, which is known as tiling, improves computational efficiency by allowing the model to oversee larger datasets during training.  As shown in Fig.\ref{fig:preprocess}, the process we follow is divided in two main steps:

\begin{itemize}
    \item The first step is to slice each aerial image and its corresponding ground truth label from the dataset. The labels are encoded in a single channel with values ranging from 0 to 4, representing background (0), building (1), woodland (2), water (3), and road (4). Aerial images have three channels (RGB). The image and label have varying heights ($H1$) and widths ($W1$) depending on the source image. Subsequently, the image and label were sliced into smaller tiles with a fixed size of $512 \times 512$ pixels ($H3$ and $W3$) without overlapping. It is important to note that the exact location of each extracted tile within the original image was preserved using metadata stored in each new tile using the rasterio library. A total of $N = 10674$ training images, derived from the original 41 aerial captures, were created through this process. 
    \item The second step is preprocessing satellite images. Each satellite raster possesses same number of bands in depth ($C2$) and have dimensions ($H2$, $W2$) that vary depending on the size of the corresponding aerial image. For each sliced aerial image, the coordinates reference system (CRS) were used to define a bounding box, which was then applied to the corresponding satellite raster to perform the cropping. This process generated $N = 10674$ smaller satellite rasters, the same number as the sliced images. These rasters retain the original number of channels ($C2$) but have a reduced size of $26 \times 26$ pixels ($H4$ and $W4$).
\end{itemize}

The normalization process was applied to the aerial images during the slicing in the pre-processing stage and to the satellite images during the acquisition phase. Regarding the later's resampling, it had a minimal impact on the satellite images due to the use of the nearest neighbor method.

\begin{figure*}[htb]
    \centering
    \includegraphics[width=\linewidth]{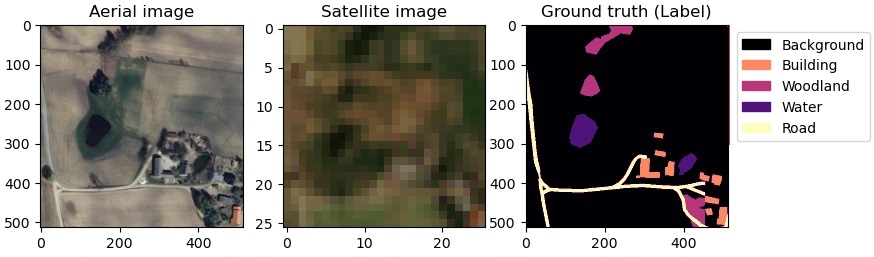}
    \caption{Difference between aerial and satellite images after preprocessing}
    \label{fig:diff_sat_ai_ap}
\end{figure*}

This preprocessing ensures that the training, validation, and test sets consistently pair corresponding inputs from both sources. Maintaining this consistent pairing is essential for effective model training and allows us to compare our results with the results of the dataset \bluecite{boguszewski_landcoverai_2022}. 
Fig.\ref{fig:diff_sat_ai_ap} shows an example of an aerial image, its corresponding label, and the corresponding satellite image.
The aerial input and the label output had a resolution of 512 pixels by 512 pixels, whereas the satellite input had a resolution of 26 pixels by 26 pixels.

As for the training process, in order to address the potential issue of class imbalance in the training data, we employed class weights during loss calculation. These weights compensate for the unequal distribution of pixels across different land cover classes. The equation Eq.\eqref{eq:Freq} defines the calculation of class weights. The frequency of class $i$ is computed as the sum of all pixels labeled as class $i$ divided by the total number of pixels in the training labels. Thus, the weight for class $i$ is calculated as the inverse of this frequency.

\begin{multline}
  \text{ClassWeight}_{i} = \frac{1}{\text{FrequencyClass}_{i}} \\
  \text{ where FrequencyClass}_{i} = \frac{\sum{\text{Pixel}_{i}}}{\sum{\text{Pixels}}}\\
  \label{eq:Freq}
  \end{multline} 

\begin{table}[htb]
    \centering
    \caption{LandCover.ai classes labels statistics on the training set.}
    \footnotesize
    \begin{tabular}{llllll}
         Class & Buildings & Woodlands & Water & Roads & Background \\\hline
         Frequency & 0.86\% & 33.14\% & 6.46\% & 1.62\% & 57.92\% \\
         Weight & 0.58794 & 0.01520 & 0.07797 & 0.31017 & 0.00869\\ 
    \end{tabular}
    \label{tab:trainStats}
\end{table}

Tab.\ref{tab:trainStats} presents statistical data derived from the 7470 labels in the training set. As illustrated by the table, the "Buildings" and "Roads" classes are underrepresented, collectively accounting for less than 3\% of the total pixels. In contrast, the "Background" class dominates the dataset, comprising 57.92\% of the pixels. These observations highlight the presence of class imbalance in the training data, hence the need to apply appropriate class weights during loss calculation. The specific weight values for each class are provided in the second row of Tab.\ref{tab:trainStats}.

As for data augmentation, even thought it is possible for the $input_1$ consisting of aerial images it is not applicable for the $input_2$ due to the later been in a smaller resolution and the values of pixels are sensitive. 
In the case of using data augmentation for comparison purposes we used the same configuration as in \bluecite{boguszewski_landcoverai_2022}, in which we also apply  the nine augmentations for each $input_1$ randomly change: sharpness, hue saturation, contrast, grayscale,  brightness, adding noise, rotation, doing flipping, padding, and cropping. Therefore, we also obtained 74700 samples in the training set. 

\section{Experiment setup}
The experiments were conducted on a local machine equipped with a dual NVIDIA RTX A5000 (each with 24GB GDDR6 memory), a Xeon(R) processor W-2223 (3.60GHz octa-core), and 32 Gb RAM, which were running a Linux system Ubuntu 22.04. 

In order to ensure the reproducibility of the experiments, a random seed of $3407$ was set, which was inspired by \bluecite{Picard_2021}. The PyTorch-Lightning 2.0.1 \bluecite{Falcon_PyTorch_Lightning_2019} was used as the backbone coding framework and the complete experiments were monitored using MLFlow\bluecite{zaharia2018accelerating}. Therefore, all experiments employed a deterministic configuration for the dual GPUs, utilizing Automatic Mixed Precision with a half floating-point format (16-bit mixed precision) to accelerate training. This configuration permitted the utilization of Distributed Data Parallel with larger batch sizes, thereby facilitating accelerated model learning. 
For the retrieval of the custom weights for the Xception Aligned pretrained backbone (XA65), we used the ones provided by HuggingFaces, made available through the TIMM library \bluecite{rw2019timm}. This approach facilitates faster training by utilizing pre-trained features that are relevant to image segmentation tasks.
In regard to the issue of overfitting, our approach involved the use of batch normalization and early stopping after 15 epochs of no improvement in the mIoU in the validation set. The application of data augmentation was not deemed appropriate due to the potential for modifying the initial values associated with the satellite input, which had a relatively low resolution of 26 by 26.

In addition to these experimental setups and optimization strategies, the DIFD model employs the Dice Cross Entropy Loss Eq.\eqref{eq:DiceCE}, which is a combination of two losses: Dice Loss (Eq.\eqref{eq:DiceLoss}) and the Cross Entropy Loss (CELoss) (Eq.\eqref{eq:CELoss}). 
The DiceLoss function is designed to enhance the model's IoU score for each class, thereby improving the accuracy of the segmentation process. This is achieved by imposing a penalty on the model for any missed or incorrectly classified pixels. In contrast, the CELoss function measures the discrepancy between the true labels and the model's predictions, incorporating the class weights (CW) values specified in Table.\ref{tab:trainStats} to address the class imbalances. The class weights serve to adjust the loss function, thereby emphasizing underrepresented classes during training. Also Optuna  is used to perform a comprehensive hyperparameter search. The optimal hyperparameters identified through this process are detailed in Table \ref{tab:HPtab4}.

\begin{table}[htb]
    \footnotesize
        \caption{Hyper parameters used for all models.}
        \centering
        \begin{tabular}{lll}
        Name                   & Value & Impact                    \\ \hline
        Batch size & 26 & Images processed in one step \\
        Max Epoch & 100 & Limit the number of epochs \\
        Initialization type     & Kaiming & Initialization for weights         \\
        Learning rate     & 0.001 & Impact learning convergence         \\
        Optimizer     & AdamW & Weights and biases update         \\
        Workers & 7 & CPUs to use for loading data \\
        \end{tabular}
        \label{tab:HPtab4}
    \end{table}

In addition to these experimental setups and optimization strategies, the DIFD model incorporates the Dice Cross Entropy Loss (Eq.\eqref{eq:DiceCE}), a composite loss function that combines Dice Loss (Eq.\eqref{eq:DiceLoss}) and Cross Entropy Loss (CELoss) (Eq.\eqref{eq:CELoss}). Dice Loss, which is frequently employed in segmentation tasks, enhances the model’s Intersection over Union (IoU) score by penalizing incorrect classifications, which improves the accuracy of the segmentation. This loss function calculates the overlap between predicted and actual segments, favoring predictions that closely match the ground truth. In contrast, Cross Entropy Loss measures the difference between the predicted class probabilities and the true class labels, effectively guiding the model to reduce classification errors. To address class imbalances, class weights (CW), as detailed in Table.\ref{tab:trainStats}, are incorporated into the CELoss function. These weights adjust the contribution of each class to the overall loss, ensuring that underrepresented classes are emphasized during training to improve the model's performance across all classes.

\begin{multline}
  \text{DiceCELoss}(y_{\text{pred}}, y_{\text{true}}) = \\
  \text{DiceLoss}(y_{\text{pred}}, y_{\text{true}}) + 
  \text{CELoss}(y_{\text{pred}}, y_{\text{true}})\\
  \label{eq:DiceCE}
\end{multline}
\begin{multline}
  \text{DiceLoss}(y_{\text{pred}}, y_{\text{true}}) = \\ 1 - \frac{2\times y_{pred}\bigcap y_{true}}{y_{pred}\bigcup y_{true}} 
  = 1- \frac{2 \times \sum_{i=1}^{N} y_{\text{true}}^{(i)} \cdot y_{\text{pred}}^{(i)}}{\sum_{i=1}^{N} y_{\text{true}}^{(i)} + \sum_{i=1}^{N} y_{\text{pred}}^{(i)}}\\
  \label{eq:DiceLoss}
\end{multline}
\begin{multline}
  \text{CELoss}(y_{\text{pred}}, y_{\text{true}}) = \\ -CW \times
  \frac{1}{N} \sum_{i=1}^{N}\left( {y_{\text{true}}^{(i)} \cdot \log(y_{\text{pred}}^{(i)}) + (1 - y_{\text{true}}^{(i)}) \cdot \log(1 - y_{\text{pred}}^{(i)})} \right) \\
  \label{eq:CELoss}
\end{multline}

The main metric for evaluation of semantic segmentation task is the IoU. While the IoU calculates the intersection over union for each class individually Eq.\eqref{eq:IoU}, the mean IoU (mIoU) calculates the average IoU for all classes.

\begin{equation}
    \begin{aligned}
        IoU =\frac{\left |y_{pred} \cap y_{true} \right|}{\left |y_{pred} \cup  y_{true} \right|} = \frac{TP}{TP + TN + FP}
    \end{aligned}
    \label{eq:IoU}
\end{equation}

In the same way, to obtain more information about the performance of the models, we used also the F1\_score Eq.\eqref{eq:F1}.


\begin{equation} \text{F1\_score} = \frac{2TP}{2TP + FP + FN} \label{eq:F1}\end{equation}


In this context, True Positives (TP) refer to the number of positive samples that are correctly classified by the model. True Negatives (TN) represent the correctly identified negative samples, while False Positives (FP) denote the number of negative samples that are incorrectly classified as positive. Finally, False Negatives (FN) account for the number of positive samples that are mistakenly classified as negative.

\section{Results and discussion}
In this section, we present the main results of DIFD\_UpConvT and compare them with various models based on the DeepLabV3+ architecture, such as the original DeepLabV3+, the enhanced version of DeepLabV3+, DIFD\_UpBilinear, DIFD\_UpNearest, and  DIFD\_UpPS using the ablation approach.

In general, the models tend to converge after 60 epochs based on the loss and the saturation on the IoU during training, with some inconstancy during the early epochs of the validation. 
This behavior is attributable to the loss function DiceCELoss used, which attempts to enhance the segmentation of each class despite the presence of imbalanced data.  
Based on our results, it can be seen that the classes building and road are the most difficult classes to learn, this particularity is related to the fact that these two classes had less than 3\% of the dataset.

To better understand the impact of various experimental configurations, we performed an ablation study on our approach. This study was organized into five major categories: using aerial input only (IDs 0, 1, and 3), satellite input only (ID 3), dual input where the aerial image was resized to match the satellite input resolution (IDs 4 and 5), dual input with both aerial and satellite data using DIFD\_UpConvT with different band selections, and dual input with aerial and satellite data using alternative upsampling methods.
To ensure a comprehensive evaluation, we tested different model setups. The Base configuration employed the original DeepLabV3+ model with hyperparameters from the study by Boguszewski et al. \bluecite{boguszewski_landcoverai_2022} (ID 0). The setup of the model that used the fine-tuned hyperparameters obtained through Optuna, that we named HP Tuned, was detailed in table Tab.\ref{tab:HPtab4}. The HP Tuned + UpConvT configuration added an $UpConvT$ block to the fine-tuned model, as illustrated in Fig. \ref{fig:DIFD_F5}. The HP Tuned + UpConvT + Dual Input setup incorporated both aerial and satellite inputs but with aerial input only. We also evaluated the HP Tuned + UpConvT + Dual Input + Augmented Data configuration, which included augmented data. Additionally, the DIFD\_UpConvT model was tested with various band combinations, while comparisons were made with DIFD\_UpBilinear, DIFD\_UpNearest, and DIFD\_UpPS, each representing different upsampling techniques as described in subsection \ref{Variation_of_DFID}. This approach helps demonstrate that the observed improvements are due to specific modifications and not merely different implementations or hyperparameter variations.

The results of our experiments, as demonstrated in Tab.\ref{tab:RtabAll}, indicate that fusing RGB images from aerial and satellite sources generally yielded higher Intersection over Union (IoU) and F1-score metrics compared to using only aerial images. However, this was not the case in scenarios involving augmented data. Notably, the DIFD\_UpConvT model utilising the $7B$  configuration attained the highest F1-score of 91.5\%. This performance is corroborated by an IoU score of 84.91\% for the same DIFD\_UpConvT model with the $7B$, configuration, thereby demonstrating superior segmentation performance relative to other configurations tested in our experiments.

\begin{table*}[htb]
\centering
\footnotesize
\caption{DIFD's design choices ablation and performance in terms of various metrics}
\begin{tabular}{lllllll}
\hline
ID & Configuration for DeepLabV3+ (OS16)    & Aerial & Satellite & Input   & mF1-Score & mIoU  \\ \hline
0  & Base                                  & \checkmark      &           & RGB     & 89.25     & 81.33 \\
1  & HP tuned                              & \checkmark      &           & RGB     & 89.95     & 82.56 \\
2  & HP tuned + UpConvT                    & \checkmark      &           & RGB     & 90.12     & 82.74 \\ \hline
3  & HP tuned + UpConvT                    &        & \checkmark         & 10B     & 14.54     & 11.43 \\ \hline
4  & HP tuned + UpConvT + Dual input       & \checkmark      &           & RGB+RGB & 90.8      & 83.84 \\
5  & HP tuned + UpConvT + Dual input + Aug & \checkmark      &           & RGB     & 92.06     & 85.78 \\ \hline
6  & DIFD\_UpConvT                         & \checkmark      & \checkmark         & RGB+4B & 90.75    & 83.7 \\
7  & DIFD\_UpConvT                         & \checkmark      & \checkmark         & RGB+10B & 90.94     & 84.07 \\
8  & DIFD\_UpConvT                         & \checkmark      & \checkmark         & RGB+7B  & 91.5      & 84.91 \\ \hline
9  & DIFD\_UpBilinear                      & \checkmark      & \checkmark         & RGB+7B  & 91.32     & 84.63 \\
10  & DIFD\_UpNearest                       & \checkmark      & \checkmark         & RGB+7B  & 91.3      & 84.56 \\
11  & DIFD\_UpPS                            & \checkmark      & \checkmark         & RGB+7B  & 91.28     & 84.53 \\\hline
\end{tabular}
\label{tab:RtabAll}
\end{table*}

In order to evaluate the segmentation score accurately, it is necessary to consider each class IoU, as presented in Tab.\ref{tab:RtabIoU}. As the DIFD\_UpConvT demonstrated the highest score in terms of mIoU (except for the configuration in which the augmentation were used), a subsequent observation indicated that it effectively distinguished between the "buildings", "woodlands", and "background" classes , with a respective IoU of 78.68\%, 90.86\%, and 92.94\% for these classes. Furthermore, the remaining classes exhibited commendable performance, nearly matching the top results and better than without the fusion and without data augmentation.

\begin{table*}[htb]
    \centering
    \caption{Intersection over Union (IoU) on the test set for each classes}
    \footnotesize 
    \begin{tabular}{llp{1.43cm}p{1.43cm}p{1.43cm}p{1.43cm}p{1.43cm}}  \hline
        ID & Input   & Buildings & Woodlands & Water & Roads & Background \\ \hline
        0  & RGB     & 71.85     & 87.3      & 93.19 & 64    & 90.35      \\
        1  & RGB     & 68.64     & 90.7      & 93.88 & 67.11 & 92.46      \\
        2  & RGB     & 75.25     & 88.9      & 93.28 & 64.36 & 91.88      \\ \hline
        3  & 10B     & 0         & 0         & 0     & 0     & 57.13      \\ \hline
        4  & RGB+RGB & 76.66     & 90.3      & 94.12 & 65.59 & 92.54      \\
        5  & RGB     & 79.11     & 91.5      & 94.51 & 70.39 & 93.39      \\ \hline
        6  & RGB+4B & 75.89     & 90.3      & 93.32 & 66.71 & 92.3 \\
        7  & RGB+10B & 75.98     & 90.8      & 94.22 & 66.54 & 92.79      \\
        8  & RGB+7B  & 78.68     & 90.9      & 94.29 & 67.79 & 92.94      \\ \hline
        9 & RGB+7B  & 77.76     & 90.7      & 94.35 & 67.55 & 92.81      \\
        10 & RGB+7B  & 77.43     & 90.2      & 94.24 & 68.49 & 92.41      \\
        11 & RGB+7B  & 78.22     & 90.4      & 93.87 & 67.51 & 92.64      \\ \hline
    \end{tabular}
    \label{tab:RtabIoU}
\end{table*}

The incorporation of the multi-spectral satellite image is likely to have contributed to the improved IoU scores observed for several classes. The image provides additional information, including NDVI, NDWI and SCL which can assist in differentiating between land cover types, particularly minority classes. 
In the original work by \bluecite{boguszewski_landcoverai_2022}, the high IoU score was primarily driven by the dominant background class. However, the performance of the road class was significantly lower. The proposed approach yielded a notable improvement in terms of IoU for the road class, demonstrating the efficacy of multi-spectral data fusion in enhancing segmentation accuracy for smaller classes.

\begin{figure*}[htb]
    \centering
    \includegraphics[width=\textwidth]{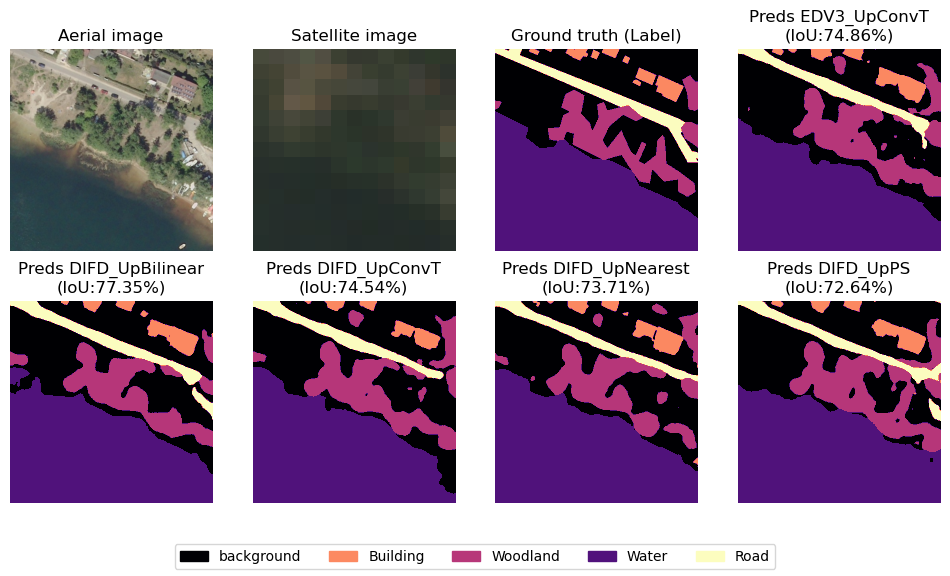}
    \caption{Example 1 for the source materials and the predictions from various models}
    \label{fig:E1}
\end{figure*}

Fig.\ref{fig:E1} presents an example from the test set, which was also used in the original paper for closer examination of the model predictions. 
The satellite raster had 7 bands, but we only plotted the RGB bands. Even thought the pixel resolution is low, each pixel had information about 40 by 40 pixel in the aerial images.
Note that we named the configuration ID 5 as EDV3\_UpConvT for samplicity. 
While the DIFD\_UpBilinear model achieved the highest IoU for this specific instance and visually appeared closest to the ground truth label, other models exhibited tendencies to over-predict woodlands at the expense of the background class and displayed lower precision for road pixels.
The particularity of this example is that the satellite image does not provide better information to the aerial image and the highest score with upsampling the raster with the bilinear method is due to the smoothing effect that was helpful on this image rather than the raw values.

\begin{figure*}[htb]
    \centering
    \includegraphics[width=\textwidth]{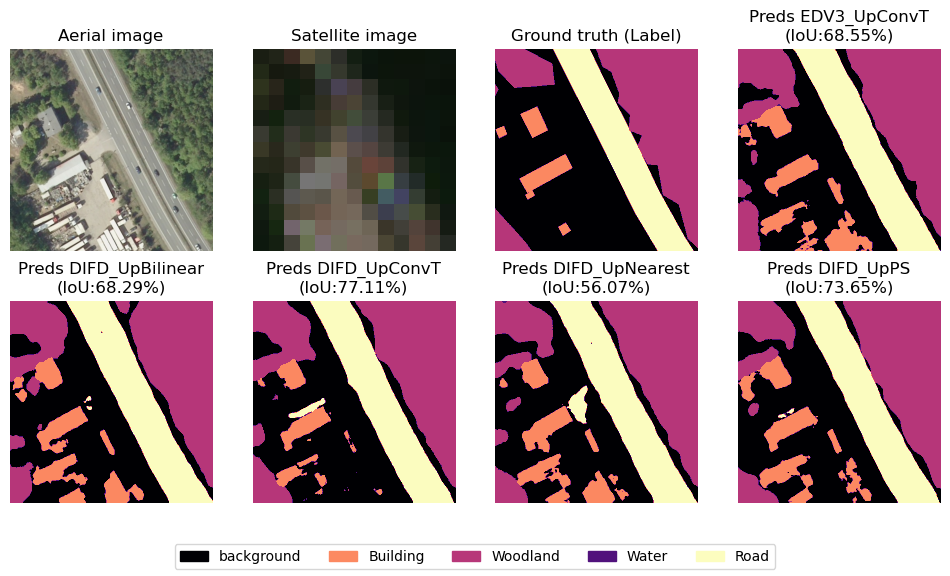}
    \caption{Example 2 for the source materials and the predictions from various models}
    \label{fig:E2}
\end{figure*}

Similarly, we showed in Fig.\ref{fig:E2}  another example that visualizes one of the main confusions noticed across all models. This is related to the building class. From the aerial image and the labels, we can see that there exist some long trucks stationed next to some buildings. Due to the similar structure of these long trucks with the building, all models confuse it to some degree to the building class instead of the background class.

Our analysis identified instances of confusion between certain classes, particularly with the background class. It would be optimal to introduce more specific land cover classes to address this issue. However, the constraints of time and resources precluded the implementation of this solution.
When combining information from aerial images and satellite data, we explored the use of nearest interpolation, which offers computational efficiency and value-preserving nature. While this method effectively retains the original data values, a closer examination reveals potential drawbacks. Due to the discrepancy in resolution between the raster data and the requisite upscaled dimensions, the UpNearest methodology merely duplicates pixels, which may result in a "grill effect" in the generated output.
An alternative approach is bilinear interpolation, which replaces pixel duplication with the smoothing of the raster input. Nevertheless, both UpNearest and UpBilinear are subject to limitations when confronted with discrepancies in resolution.

\section{Conclusion}

This paper presents an enhanced DeepLabV3+ architecture for land cover segmentation that leverages the combined power of aerial images and satellite data. We have proposed a novel decoder integration approach to effectively fuse information extracted from these diverse remote sensing data sources. The model was evaluated on the LandCover.ai dataset (aerial) and Sentinel-2 data (satellite), achieving a promising mean Intersection over Union (mIoU) of 84.91\% without data augmentation. This result significantly outperforms existing methods without the use of augmented data. A key contribution of our work lies in using satellite data and modifying the upsampling process within the DeepLabV3+ architecture. We replaced standard bilinear upsampling with a custom weighted upsampling module utilizing deconvolution layers. The modification demonstrably improves segmentation accuracy, as confirmed by our experimental results. While the proposed method successfully exploits the complementary information from aerial and satellite imagery, there exist opportunities for further exploration. The model's accuracy might benefit from improvements in capturing intricate land cover features in complex landscapes. Additionally, relying on specific datasets like LandCover.ai and Sentinel-2 may introduce limitations in generalizability. Investigating the model's performance across a wider range of datasets is an important future direction. Furthermore, the exploration of alternative fusion strategies has the potential to enhance segmentation accuracy further. This could involve the integration of additional remote sensing modalities or the incorporation of contextual information into the model.

\section*{Acknowledgements}
This project has been funded by the Ministry of Europe and Foreign Affairs (MEAE), the Ministry of Higher Education, Research (MESR) and  the Ministry of Higher Education, Scientific Research and Innovation (MESRSI), under the framework of the Franco-Moroccan bilateral program PHC TOUBKAL Toubkal/21/121 2023, with Grant number: 45942UG.  

\bibliographystyle{IEEEtran} 
\bibliography{Ref}
\end{document}